\documentclass[10pt,twocolumn,letterpaper]{article}

\usepackage{cust}
\usepackage{times}
\usepackage{epsfig}
\usepackage{graphicx}
\usepackage{amsmath}
\usepackage{multirow}
\usepackage{amssymb}
\usepackage{xcolor}
\usepackage{enumitem}

\usepackage{subfig}
\usepackage[pagebackref=true,breaklinks=true,letterpaper=true,colorlinks,bookmarks=false]{hyperref}

\begin{document}

% ---------------------------------------------------------------------------------------------- %
%% TITLE & AUTHORS
% ---------------------------------------------------------------------------------------------- %
\title{VideoMem: Constructing, Analyzing, Predicting Short-term and Long-term Video Memorability}

%%first author
\author{Romain Cohendet\\
Technicolor\\
975 Avenue des Champs Blancs, 35576 Cesson-Sevigne, France\\
{\tt\small romain.cohendet@technicolor.com}
%%second author
\and
Claire-H\'el\`ene Demarty\\
%\\
%\\
{\tt\small claire-helene.demarty@technicolor.com}
%%third author
\and
Ngoc Q. K. Duong\\
%\\
%\\
{\tt\small quang-khanh-ngoc.duong@technicolor.com}
%%fourth author
\and
Martin Engilberge\\
%\\
%\\
{\tt\small martin.engilberge@technicolor.com}
}

\maketitle

\begin{abstract}
Humans share a strong tendency to memorize/forget some of the visual information they encounter. This paper focuses on providing computational models for the prediction of the \textit{intrinsic} memorability of visual content.
To address this new challenge, we introduce a large scale dataset (VideoMem) composed of 10,000 videos annotated with memorability scores.
In contrast to previous work on image memorability -- where memorability was measured a few minutes after memorization -- memory performance is measured twice: a few minutes after memorization and again 24-72 hours later.
Hence, the dataset comes with short-term and long-term memorability annotations.
After an in-depth analysis of the dataset, we investigate several deep neural network based models for the prediction of video memorability.
Our best model using a ranking loss achieves a Spearman's rank correlation of 0.494 for short-term memorability prediction, while our proposed model with attention mechanism provides insights of what makes a content memorable.
The VideoMem dataset with pre-extracted features is publicly available\footnote{\url{https://www.technicolor.com/dream/research-innovation/video-memorability-dataset }}. 
\end{abstract}

\section{Introduction}
\label{sec:intro}

%%paragraph 1
While some contents have the power to burn themselves into our memories for a long time, others are quickly forgotten \cite{isola_2011_makes}.
Evolution made our brain efficient to remember only the information relevant for our survival, reproduction, happiness, \emph{etc}.
This explains why, as humans, we share a strong tendency to memorize/forget the same images, which translates into a high human consistency in image memorability (IM) \cite{khosla_2015_understanding}, and probably also a high consistency for video memorability (VM).
This \textit{shared-across-observers} part of the memorability is the most obvious one to capture by machines, as it can be assessed by averaging individual memory performances, avoiding to deal with individual differences.
It also has a very broad range of applications in various areas, including education and learning, content retrieval and search, content summarizing, storytelling, content filtering, \emph{etc}. For these reasons, this paper targets the prediction of the part of the memorability that is shared by humans.

The study of VM from a computer vision point of view is a new field of research, encouraged by the success of IM, which has attracted increasing attention since the seminal work of Isola \textit{et al.} \cite{isola_2011_makes}.
In contrast to other cues of video importance, such as aesthetics, interestingness or emotions, memorability has the advantage of being clearly definable and objectively measurable (\emph{i.e.,} using a measure that is not influenced by the observer's personal judgment). This certainly participates to the growing interest for its study.
IM has initially been defined as the probability for an image to be recognized a few minutes after a single view, when presented amidst a stream of images \cite{isola_2011_makes}.
This definition has been widely accepted within subsequent work \cite{mancas_2013_memorability,kim_2013_relative,celikkale_2013_visual,khosla_2015_understanding,lahrache_2016_bag}).

%paragraph 3: from IM to VM prediction
The introduction of deep learning to address the challenge of IM prediction causes models to achieve results close to human consistency \cite{khosla_2015_understanding,baveye_2016_deep,zarezadeh_2017_image,jing_2017_predicting,squalli_2018_deep,fajtl_2018_amnet}.
As a result of this success, researchers have recently extended this challenge to videos \cite{han_2015_learning,shekhar_2017_show,cohendet_2018_annotating,cohendet_2018_mediaeval}.
However, this new research field is nascent.
%paragraph 4: need for a large database
As argued in \cite{cohendet_2018_annotating}, releasing a large-scale dataset for VM would highly contribute to launch this research field, as it was the case for the two important dataset releases in IM \cite{isola_2011_makes,khosla_2015_understanding}.
%paragraph 5: long-term memorability & passage of time
Such a dataset should try to overcome the weaknesses of the previously released datasets.
In particular, previous research on IM focused on the measurement of memory performances only a few minutes after memorization.
However, passage of time is a factor well-studied in psychology for its influence on memory, while having been largely ignored by previous work on IM, probably because of the difficulty to collect long-term memorability annotations at a large scale, in comparison with short-term ones.
Measuring a memory performance a few minutes after the encoding step is already a measure a long-term memory, since short-term memory usually lasts less than a minute for unrehearsed information\cite{revlin_cognition:_2012}.
However, memories continue to change over time: going through a consolidation process (\emph{i.e.,} the time-dependent process that creates our lasting memories), some memories are consolidated and others are not \cite{mcgaugh_2000_memorycentury}.
In other words, as claimed in \cite{cohendet_2016_prediction}, short-term memory performances might be poor predictors of longer term memory performances.
%paragraph 6: two measures of memorability
Since long-term memorability is more costly and difficult to collect than short-term memorability, it would nevertheless be interesting to know if the former can be inferred from the latter, which would also push forward our understanding of what makes a video \textit{durably} memorable. A way to achieve this consists in measuring memorability for the same videos at two points of time.
%paragraph 7: when to measure memory again?
These two measures would be particularly interesting if spaced by a time interval in which forgetting is quite significant, to maximize the size of the potentially observable differences depending on the different video features.
Observing the different forgetting curves in long-term memory (\emph{e.g.,} Ebbinghaus’s seminal work \cite{ebbinghaus_memory:_1913}), one can observe that the drop in long-term memory performance in recall follows an exponential decay and is particularly strong in the first hour, and to a lesser extent in the first day, immediately after the memorization.
Measuring long-term memory a few minutes after encoding (as done in studies of IM \cite{isola_2011_makes,khosla_2015_understanding}), and again one day or more after (\emph{i.e.,} to obtain a measure close to very long-term memory), sounds therefore a good trade-off.

%Paragraph 8: Contributions of the present study
The main contributions of this work are fourfold:
\begin{itemize}[noitemsep]
    
    \item We introduce a new protocol to objectively measure human memory of videos at two points of time (a few minutes after memorization, and 24-72 hours later) and release VideoMem, the premier large-scale dataset for VM, composed of 10,000 videos with short-term and long-term memorability scores (Sections \ref{subsec:collecting_videos} and \ref{subsec:protocol}).
	\item Through an analysis of the dataset, we address the problem of understanding VM, by highlighting some factors involved in VM (Section \ref{sec:understanding_memorability}). 
	\item We investigate three DNN-based models for VM prediction that we compare to two baseline IM models (Section \ref{sec:predicting_memorability}). The best model reaches a performance of 0.494 for Spearman's rank correlation on VideoMem.
	\item We propose an extension of the best performing model with an attention mechanism to localize what in an image makes it memorable (Section \ref{subsec:attention_mechanism}).
\end{itemize}

% ---------------------------------------------------------------------------------------------- %
%% RELATED WORK
% ---------------------------------------------------------------------------------------------- %
\section{Related work}
\label{sec:related_work}

% ================================================================================================================================================== %
%work in psychology
If long-term memory has been studied for over a century in psychology, since the seminal experimental studies of Ebbinghaus \cite{ebbinghaus_1913_memory}, its study from a computer vision point of view is quite recent, starting with \cite{isola_2011_makes}.
Images and videos had long been used as material to assess memory performances \cite{standing_1973_learning,brady_2008_visual,furman_2007_they}, proving that human posses an extensive long-term visual memory.
The knowledge accumulated in psychology helped to measure memory using classical memory tests (see \cite{richardson_1988_measures} for an extensive overview) such as recognition tests \cite{isola_2011_makes,khosla_2015_understanding,han_2015_learning,cohendet_2018_annotating} or textual question-based recall surveys \cite{shekhar_2017_show}.
Several factors are highlighted in the psychological literature for their critical influence on long-term memory, including emotion \cite{kensinger_2008_memory}, attention \cite{cowan_1998_attention}, semantics \cite{quillan_1966_semantic}, several demographic factors \cite{cohendet_2016_using}, memory re-evocation \cite{nadel_1997_memory}, or passage of time \cite{mcgaugh_2000_memorycentury}, also providing computer vision researchers with insights to craft valuable computational features for IM and VM prediction \cite{mancas_2013_memorability,isola_2014_makes,cohendet_2018_annotating}.

%IM
Focusing on the work on IM in computer vision, most studies made use of one of the two available large datasets, specifically designed for IM prediction, where IM was measured a few minutes after memorization \cite{isola_2011_makes,khosla_2015_understanding}, and consequently focused on predicting a so-called short-term IM \cite{mancas_2013_memorability,kim_2013_relative,celikkale_2013_visual,khosla_2015_understanding,baveye_2016_deep,lahrache_2016_bag,squalli_2018_deep,fajtl_2018_amnet}.
The pioneering work of \cite{isola_2011_makes} focused primarily on building computational models to predict IM from low-level visual features \cite{isola_2011_makes}, and showed that IM can be predicted to a certain extent.
Several characteristics have also been found to be relevant for predicting memorability in subsequent work, for example saliency \cite{mancas_2013_memorability}, interestingness and aesthetics \cite{isola_2014_makes}, or emotions \cite{khosla_2015_understanding}.
The best results were finally obtained by using fine-tuned or pre-extracted deep features, which outperformed all other features  \cite{khosla_2015_understanding,baveye_2016_deep,squalli_2018_deep,fajtl_2018_amnet}, with models achieving a Spearman's rank correlation near human consistency (\emph{i.e.,} $.68$) when measured for the ground truth collected in \cite{isola_2011_makes,khosla_2015_understanding}.

Work on VM is more recent. To the best of our knowledge, there exist only three previous attempts at measuring it \cite{han_2015_learning,shekhar_2017_show,cohendet_2018_annotating}.
Inspired by \cite{isola_2011_makes}, Han \textit{et al.} built a similar but far much heavier protocol to measure VM. Indeed, the long time span of the experiment makes the generalization of this protocol difficult, in particular if one targets the construction of an extensive dataset.
Another earlier approach uses questions instead of a classic visual recognition task to measure VM \cite{shekhar_2017_show}.
As a results, memorability annotations collected for the videos may reflect not only the differences in memory performances but also the differences between the questions in terms of difficulty, especially since the authors use the response time to calculate memorability scores, which might also critically depend on the complexity of the questions.
The most recent attempt at measuring VM, and the only one, to our knowledge, resulting in a publicly available dataset, comes from \cite{cohendet_2018_annotating}.
The authors introduced a novel protocol to measure memory performance after a significant retention period -- that is, weeks to years after memorization -- without needing a longitudinal study.
In contrast with previous work, the annotators did not pass through a learning task, which was replaced with a questionnaire designed to collect information about the participants' prior memory of Hollywood-like movies.
However, such a protocol implies a limited choice of content: authors needed contents broadly disseminated among the population surveyed, as the participants should have seen some of them before the task (hence the Hollywood-like movies), conducting to a number of annotations biased towards most famous content.
Furthermore, the absence of control of the memorizing process and the answers of the questionnaire based on subjective judgments make the measure of memory performance not fully objective.
To sum up, none of the previous approaches to measure VM is adapted to build a large-scale dataset with a ground truth based on objective measures of memory performance.

%VM prediction
Results obtained for VM prediction are yet far from those obtained in IM prediction.
Han \textit{et al.} proposed a method which combines audio-visual and fMRI-derived features supposedly conveying part of the brain activity when memorizing videos, which in the end enables to predict VM without the use of fMRI scans \cite{han_2015_learning}.
However, the method would be difficult to generalize.
Shekhar \textit{et al.} investigated several features, including C3D, semantic features obtained from some video captioning process, saliency features, dense trajectories, and color features, before building their memorability predictor \cite{shekhar_2017_show}.
They found that the most predictive feature combination used captioning features, dense trajectories, saliency and color features.

% ---------------------------------------------------------------------------------------------- %
%% DATASET
% ---------------------------------------------------------------------------------------------- %
\section{VideoMem: large-scale video memorability dataset}
\label{sec:dataset}
In section \ref{subsec:collecting_videos}, we describe the collection of source videos that compose the VideoMem dataset. We then introduce a new protocol to collect short-term and long-term memorability annotations for videos (Section \ref{subsec:protocol}), before explaining the computation of VM scores (Section \ref{subsec:mem_scores_calculation}).

% ================================================================================================================================================= %
\subsection{Video collection}
\label{subsec:collecting_videos}
\begin{figure}[!htbp]
	\centering
	\includegraphics[width=1\columnwidth]{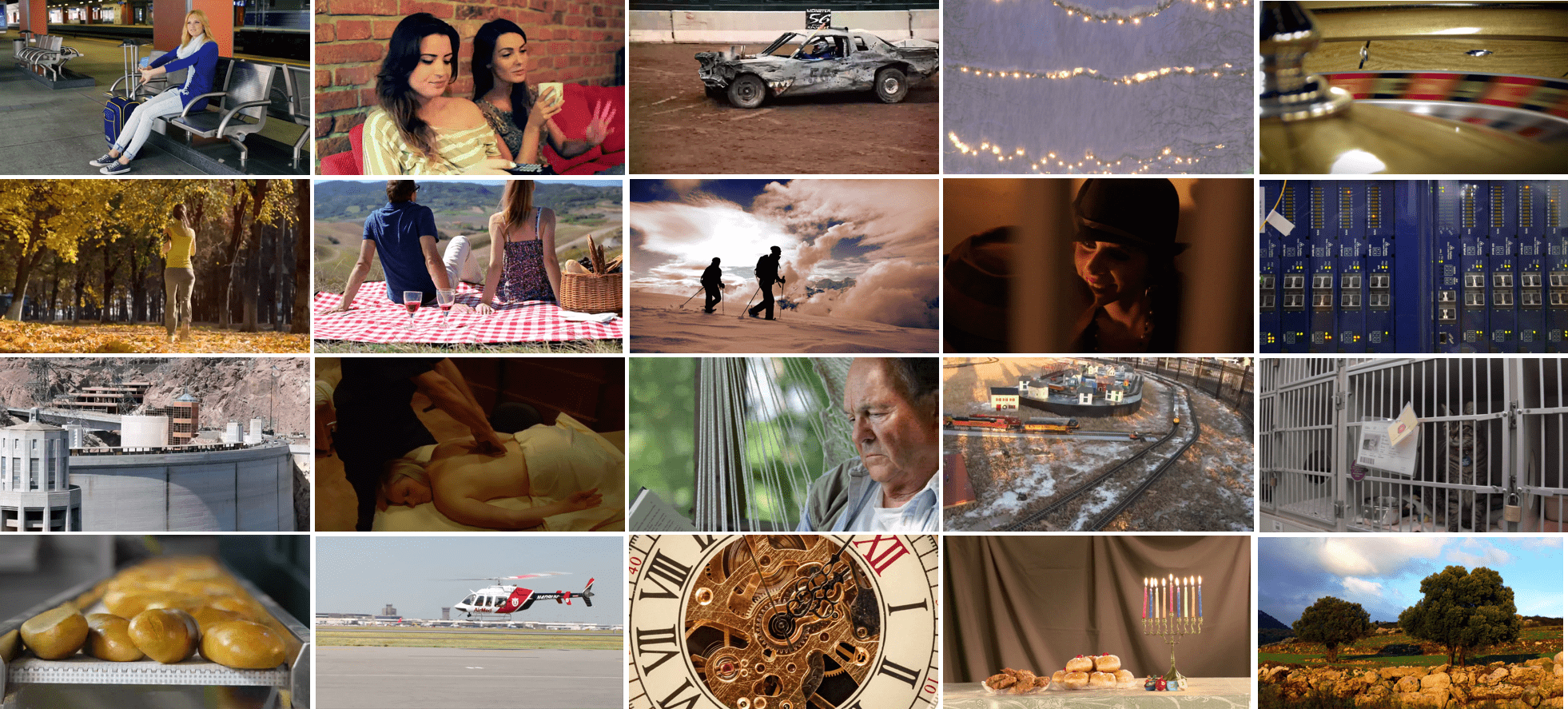}
	\caption{\label{fig:sample_videos}Example keyframes extracted from the first 20 videos (according to their names) of VideoMem, sorted by their long-term memorability (decreasing from left to right).}
\end{figure}

The dataset is composed of 10,000 soundless videos of 7~seconds shared under a license that allows their use and redistribution for research purpose only.
%neutral videos
In contrast to previous work on VM, where videos came from TRECVID \cite{shekhar_2017_show,han_2015_learning} or were extracted from Hollywood-like movies \cite{cohendet_2018_annotating}, videos in our dataset were extracted from raw footage used by professionals when creating content.
Raw footage is raw material dedicated to be further edited and included into a new motion picture, video clip, television show, movie parts, advertisements, etc.
Because such video footage is typically used to save shooting new material, it is usually generic enough to be easily integrated in different sorts of creations.
As such, they are context-independent and contain only one semantic scene. By this choice of content, we expect these basic building units to be relevant to train models which generalize on other types of videos. We are also confident that observers never saw the videos before participating in the experiment.
Videos are varied and contain different scene types such as animal, food and beverages, nature, people, transportation, etc.
A few of them contain similarities, \emph{e.g.,} same actor, same place but slightly different action, as it is the case in everyday video consumption.
A small fraction is also slow-motion. 
Each video comes with its original title, that can often be seen as a list of tags (textual metadata). Some examples of keyframes extracted from these videos are shown in Fig. \ref{fig:sample_videos}.

The original videos are of high quality (HD or 4k) and of various durations (from seconds to minutes).
As it will be described in Section~\ref{subsec:protocol}, our protocol relies on crowdsourcing. For the sake of fluency during the annotation collection and consistency between the videos, we rescaled the videos to HD and re-encoded them in \textit{.webm} format, with a bitrate of 3,000 kbps for 24 fps. 
To satisfy to the protocol's constraints, we also cut the videos to keep only the 7 seconds that represented the best the videos. 
Videos are soundless, firstly because a large part of the original data came without audio, and secondly, because it is difficult to control the audio modality in crowdsourcing.
Accordingly, memorability would be linked only to the visualization of a semantic unit, which sounds a reasonable step forward for VM prediction, without adding a potentially biasing dimension.

% ===================================================================================================================================================== %
\subsection{Annotation protocol}
\label{subsec:protocol}
To collect memorability annotations, we introduced a new protocol which enables to measure both human short-term and long-term memory performances for videos. Inspired by what was proposed in \cite{isola_2014_makes, isola_2011_makes} for IM, we also used recognition tests for our memorability scores to reflect objective measures of memory performance.
However, our protocol differs in several ways, not mentioning the fact that it is dedicated to video content.
Firstly, as videos have an inherent duration compared to images, we had to revise 1/ the delay between the memorization of a video and its recognition test and 2/ the number of videos, for the task not be too easy.
Secondly, in contrast to previous work on IM prediction, where memorability was measured only a few minutes after memorization, memory performance is measured twice to collect both short-term and long-term memorability annotations: a few minutes after memorization and again (on different items) 24-72 hours later.
The retention interval between memorization and measure is not as important as in \cite{cohendet_2018_annotating}, where it lasts weeks to years.
As previously explained, we hope, however, that this measure reflects very-long term memory performance better than short-term memorability, as forgetting happens to a large extent during the first day following the memorization.

\begin{figure*}[!htbp]
	\centering
	\subfloat[Step \#1. Interlaced encoding and recognition tasks.]{\includegraphics[width=0.8\textwidth]{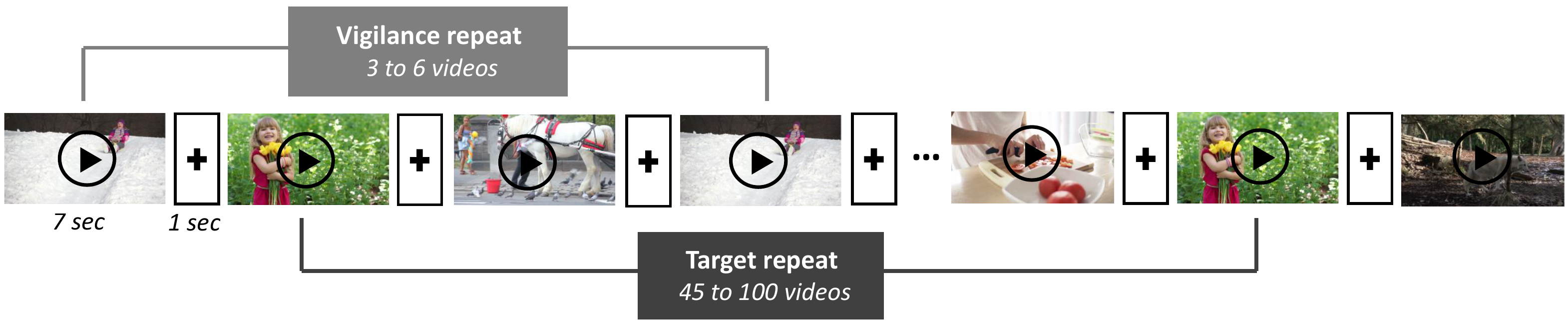}}
	\\
	\subfloat[Step \#2. Second recognition task after 24 to 72 hours.]{\includegraphics[width=0.85\textwidth]{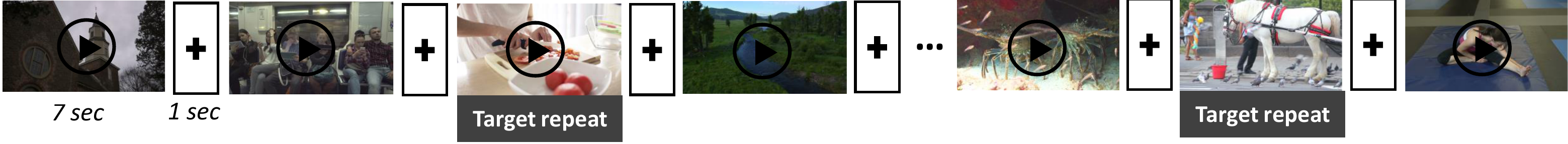}}
	\caption{\label{fig:protocol}Proposed protocol to collect both short-term and long-term video memorability annotations. The second recognition task measures memory of videos viewed as fillers during step \#1, to collect long-term memorability annotations.}
\end{figure*}

%task
Our protocol, that works in two steps, is illustrated in Fig. \ref{fig:protocol}. Step \#1, intended to collect short-term memorability annotations, consists of interlaced viewing and recognition tasks.
Participants watch a series of videos, some of them -- the \textit{targets} -- repeated after a few minutes.
Their task is to press the space bar whenever they recognize a video. Once the space bar is pressed, the next video is displayed, otherwise current video goes on up to its end.
Each participant watches 180 videos, that contain 40 \textit{targets}, repeated once for memory testing, and 80 \textit{fillers} (\emph{i.e.,} non target videos), 20 of which (so-called \textit{vigilance fillers}) are also repeated quickly after their first occurrence to monitor the participant's attention to the task. The 120 videos (not counting the repetitions) that participate to step \#1 are randomly selected among the 1000 videos that received less annotations at the time of the selection. Their order of presentation is randomly generated by following the given rule: the repetition of a \textit{target} (respectively a \textit{vigilance filler}) occurs randomly 45 to 100 (resp. 3 to 6) videos after the \textit{target} (resp. \textit{vigilance filler}) first occurrence.
%second part
In the second step of the experiment, that takes place 24 to 72 hours after step \#1, the same participants are proposed another similar recognition task, intended to collect long-term annotations.
They watch a new sequence of 120 videos, composed of 80 \textit{fillers} (randomly chosen totally new videos) and 40 \textit{targets}, randomly selected from the \textit{non-vigilance fillers} of step \#1.
Again, their task is to recognize these new targets.

%controls
Apart from the vigilance task (step \#1 only), we added several controls, settled up upon the results on an in-lab test: a minimum correct recognition rate ($15\%$, step \#2 only), a maximum false alarm rate ($30\%$ for step \#1; $40\%$ for step \#2) and a false alarm rate lower than the recognition rate (step \#2 only). This allows to obtain quality annotations by validating each user's participation; a participant could participate only once to the study.
%% PROCEDURE
We recruited participants from diverse countries and origins via the Amazon Mechanical Turk (AMT) crowdsourcing platform.
Thanks to our own controls, we did not rely on any selection of (master) workers as proposed by AMT. %to ensure the quality of the task.

%figure: experimental protocol

% ================================================================================================================================================== %
\subsection{Memorability score calculation}
\label{subsec:mem_scores_calculation}

%nb of points
After a filtering of the participants to keep only those that passed the vigilance controls, we computed the final memorability scores on 9,402 participants for short-term, and 3,246 participants for long-term memorability.
On average, a video was viewed as a repeated target 38 times (and at least 30 times) for the short-term task, and 13 times (at least 9 times) for the long-term task (this difference is inherent to the lower number of participants in step \#2). 
We assigned a first raw memorability score to each video, defined as the percentage of correct recognitions by participants, for both short-term and long-term memorability.

\begin{figure}[!htbp]
	\centering
	\subfloat[Step \#1. Recognition rate decreases linearly over time.]{\includegraphics[width=0.47\columnwidth]{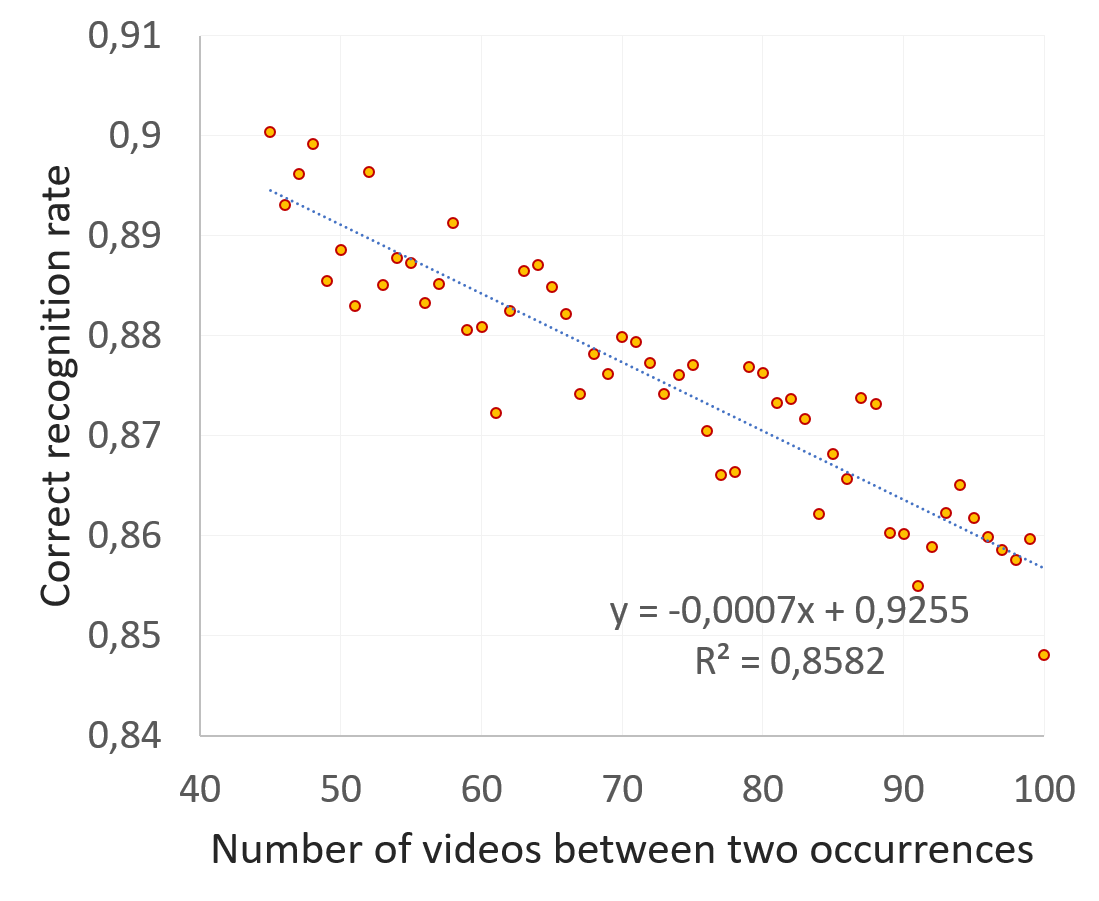}}
	\quad
	\subfloat[Memory performances does not significantly change between 24 and 72 hours after memorization.]{\includegraphics[width=0.47\columnwidth]{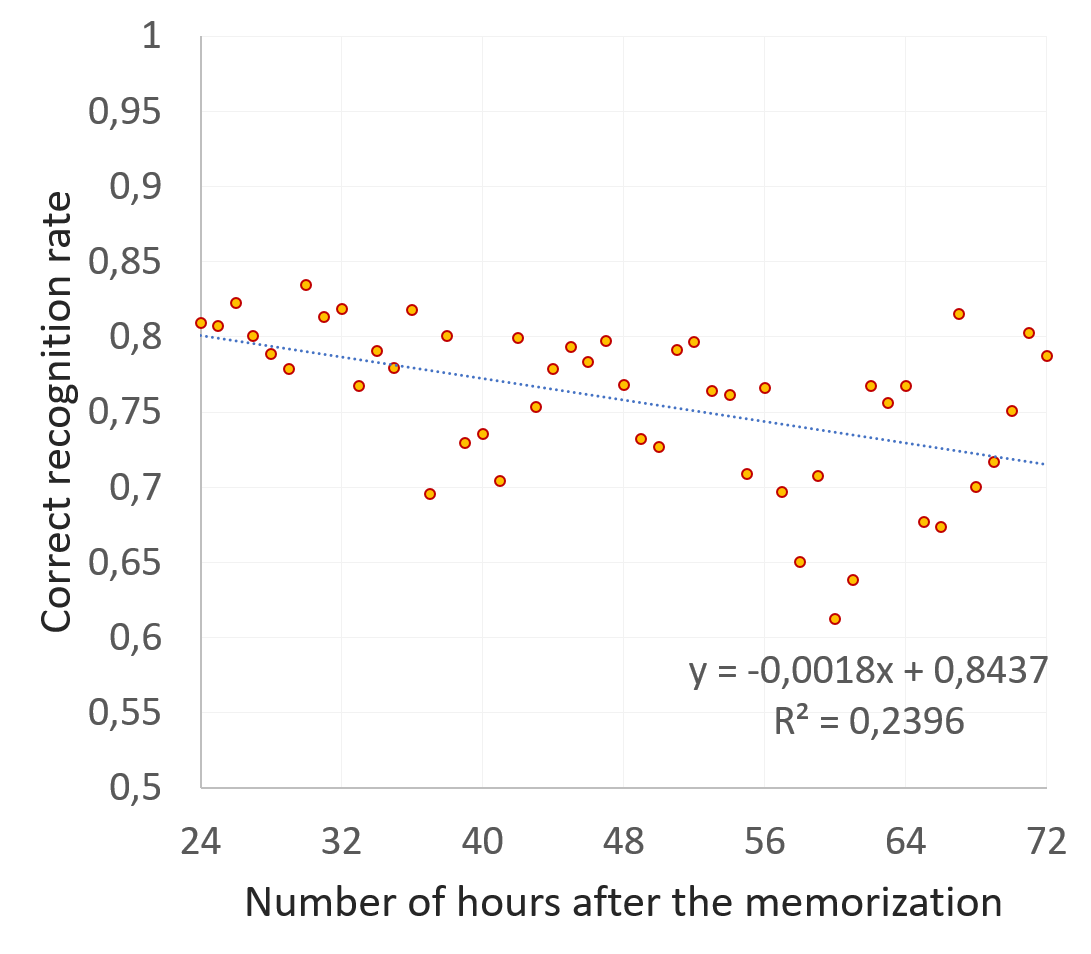}}
	\caption{\label{fig:correction}Mean correct recognition rate \textit{vs.} the retention interval between the memorization and the measure of memory performance. Blue lines represent linear fitting.}
\end{figure}

%Memorability correction
The short-term raw scores are further refined by applying a linear transformation that takes into account the memory retention duration to correct the scores.
Indeed, in our protocol, the repetition of a video happens after variable time intervals, \emph{i.e.,} after 45 to 100 videos for a \textit{target}.
In \cite{isola_2014_makes}, using a similar approach for images, it has been shown that memorability scores evolve as a function of the time interval between repeats while memorability ranks are largely conserved.
We were able to prove the same relation for videos, \emph{i.e.,} memorability decreases linearly when the retention duration increases (see Fig. \ref{fig:correction}, left).
Thus, as in \cite{khosla_2015_understanding}, we use this information to apply a linear correction (shown in Fig. \ref{fig:correction}) to our raw memorability scores to explicitly account for the difference in interval lengths, with the objective for our short-term memorability scores to be the most representative of the typical memory performance after the maximal interval (\emph{i.e.,} 100 videos). 
Note that the applied correction has nevertheless little effect on the  scores both in terms of absolute and relative values.
Note also that we did not apply any correction for long-term memorability scores (Fig. \ref{fig:correction}, right). Indeed, we observed no specific, strong enough relationship between retention duration and long-term memorability. % from our collected scores.
This was somehow expected from what can be found in the literature : according to our protocol, the second measure was carried out 24 to 72 hours after the first measure. After such a long retention duration, it is expected that the memory performance is no more subjected to substantial decrease due to the retention duration.
In the end, the average short-term memorability score is $0.859$ (instead of $0.875$) and the average long-term memorability score is $0.778$, all values showing a bias towards high values.

%Figure: Correction of short-term memorability.

% ================================================================================================================================================== %
% ================================================================================================================================================== %
% ---------------------------------------------------------------------------------------------- %
%% UNDERSTANDING VIDEO MEMORABILITY
% ---------------------------------------------------------------------------------------------- %
\section{Understanding video memorability}
\label{sec:understanding_memorability}

%This section gives some insights on the two concepts of short-term memorability and long-term memorability and their potential correlations.

% ================================================================================================================================================== %
\subsection{Human consistency \textit{vs.} annotation consistency}
\label{subsec:consistency}

%global consistency calculation
Following the method proposed in \cite{isola_2014_makes}, we measured human consistency when assessing VM. For this purpose, we randomly split our participants into two groups of equal size, \emph{i.e.,} 4,701 for short-term memorability, 1,623 for long-term memorability, and computed VM scores independently in each group as described in Section \ref{subsec:mem_scores_calculation}. We then calculated a Spearman's rank correlation between the two groups of scores.
Averaging over 25 random half-split trials, an average Spearman's rank correlation, \emph{i.e.,} a global human consistency, of $0.481$ is observed for short-term memorability and of $0.192$ for long-term memorability.

\begin{figure}[!htbp]
	\centering
	\includegraphics[width=0.47\columnwidth]{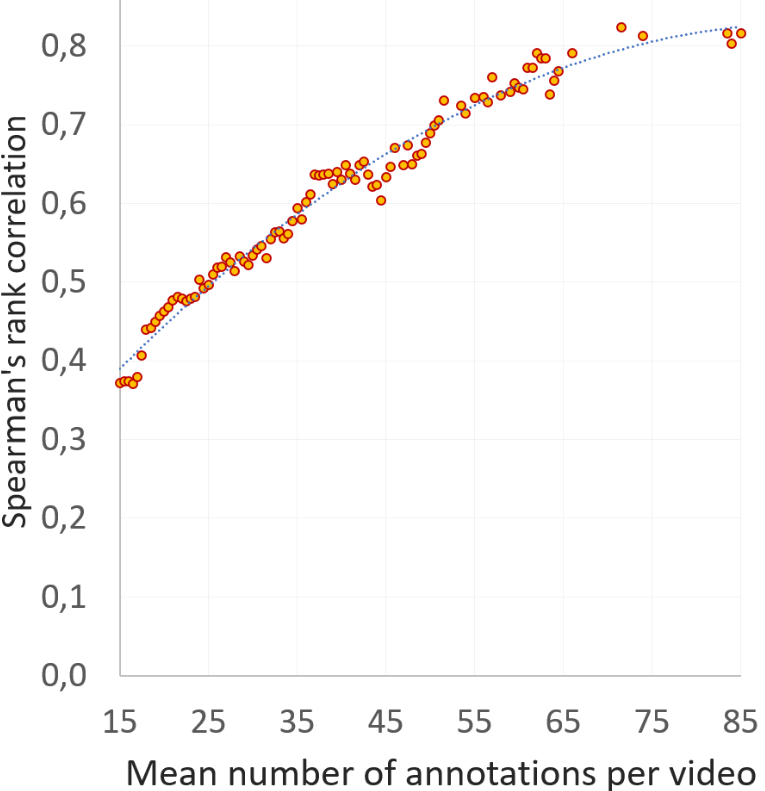}
	\includegraphics[width=0.47\columnwidth]{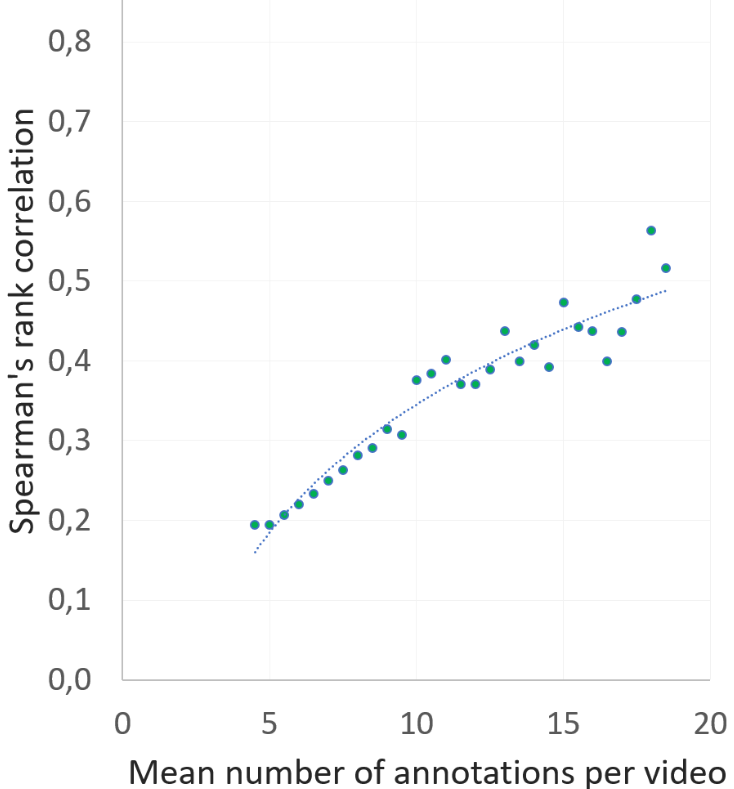}
	\caption{\label{fig:annotation_consistency}Annotation consistency \textit{vs.} mean number of annotations per video (left: short-term, right: long-term).}
\end{figure}

%Annotation consistency
Such a method divides the number of annotations that is taken into account for the score computation at least by a factor of 2. Moreover, it may ends with groups with unbalanced number of annotations per video as the split is randomly applied on the participants, not taking into account which videos they watched. For this reason, we computed a new metric so-called \textit{annotation consistency}. We reproduced the previous process of human consistency computation but on successive subparts of the dataset by considering for each sub-part only videos which received at least N annotations. Each subpart is then split in two groups of participants while ensuring a balance number of participants per video. By doing so, we obtain the annotation consistency as a function of the number of annotations per video, as presented in Fig. \ref{fig:annotation_consistency}.
This allows us to interpolate the following values: Annotation consistency reaches $0.616$ (respectively $0.364$) for the short-term (resp. long-term) task, for a number of annotations of 38 (resp. 13).
The value of $0.616$ for short-term memorability is to be compared to the one found in \cite{khosla_2015_understanding} ($0.68$) for images. Slightly lower than the latter, one should note that this consistency on VideoMem was obtained with less annotations than in the work of \cite{khosla_2015_understanding}, which is consistent with \cite{cohendet_2018_annotating}. The maximum consistency is also slightly higher for VM than for IM ($0.81$ against $0.75$ in \cite{isola_2011_makes} and $0.68$ in \cite{khosla_2015_understanding}).
An explanation is that videos contain more information than images and thus are more easily remembered
However, one should keep in mind that the protocols to collect annotations differ in several ways, making these results not fully comparable.
From Fig. \ref{fig:annotation_consistency}, we see that long-term consistency follows the same evolution as short-term consistency.

% ================================================================================================================================================== %
\subsection{Memorability consistency over time}
\label{subsec:over_time}

\begin{figure}[!htbp]
	\centering
	\includegraphics[width=0.42\columnwidth]{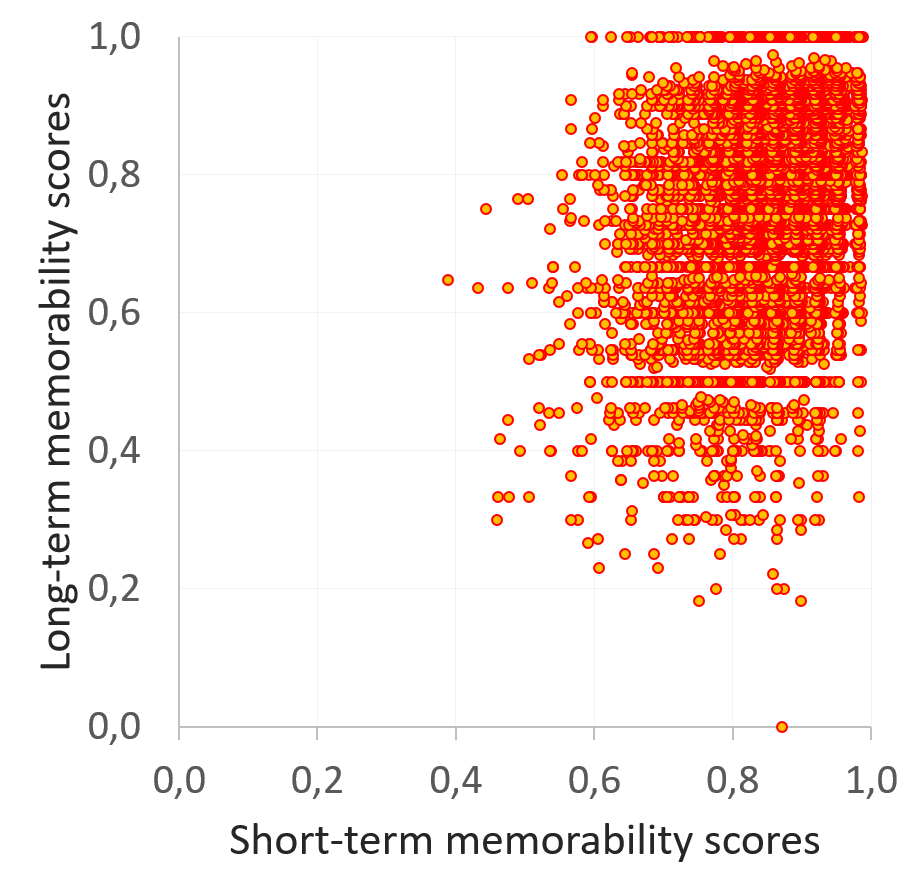}
	\includegraphics[width=0.42\columnwidth]{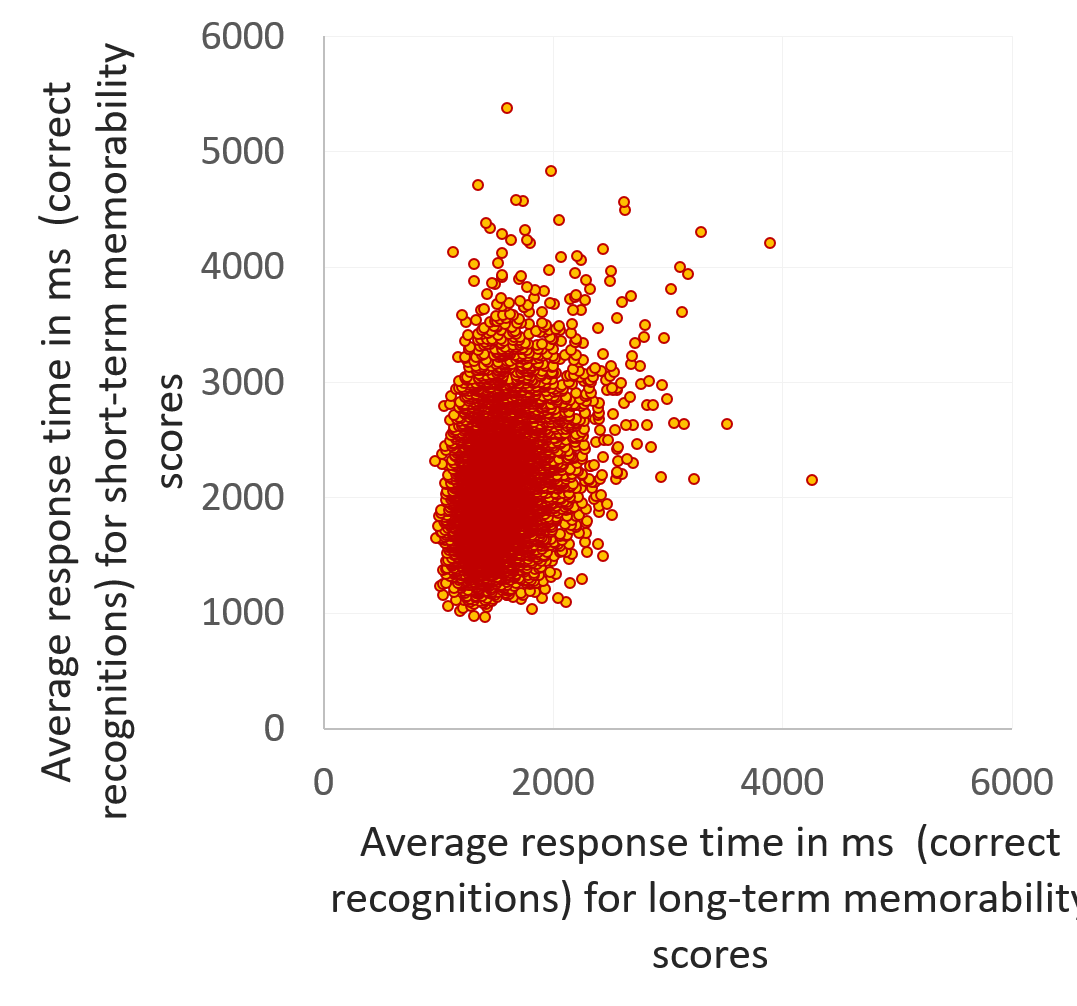}
	\caption{Short-term \textit{vs.} long-term memorability scores (left) and  average response times (right).}
	\label{fig:scatter_plot_st_vs_lt}
\end{figure}

In this study, we are interested in assessing how well memorability scores remain consistent over time, \emph{i.e.,} if a video highly memorable after a few minutes of retention remains also highly memorable after 24 to 72 hours. The computation of a Spearman's rank correlation coefficient between the long-term and short-term memorability scores for the 10,000 videos exhibits a moderate positive correlation ($\rho=0.305, p<.0001$) between the two variables, as also shown in Fig. \ref{fig:scatter_plot_st_vs_lt}.
To discard a potential bias that would come from the highest number of annotations in step \#1 compared to step \#2, we computed the correlation for the 500 most annotated videos in the long-term task (that have at least 21 annotations) and then again for the 100 most annotated (at least 28 annotations), observing similar Spearman values of $\rho=0.333$, $p<.0001$ and $\rho=0.303$, $p<.0001$, respectively.
This result suggests that memory evolves with time and in a non homogeneous manner depending on the videos: a video highly memorable a few minutes after visualization might not remain highly memorable in long-term memory. This finding is consistent with the hypothesis we proposed in the introductory section, that the information important for a content to be memorized might not be the same for short-term and long-term memorization. 

%Scatter plot: st vs. lt memorability
%\textcolor{red}{font size in axes is not consistent, and the figure seems to be quite small...}

% ================================================================================================================================================== %
\subsection{Memorability and response time}
\label{subsec:resp_time}
\begin{figure}[!htbp]
	\centering
	\includegraphics[width=0.42\columnwidth]{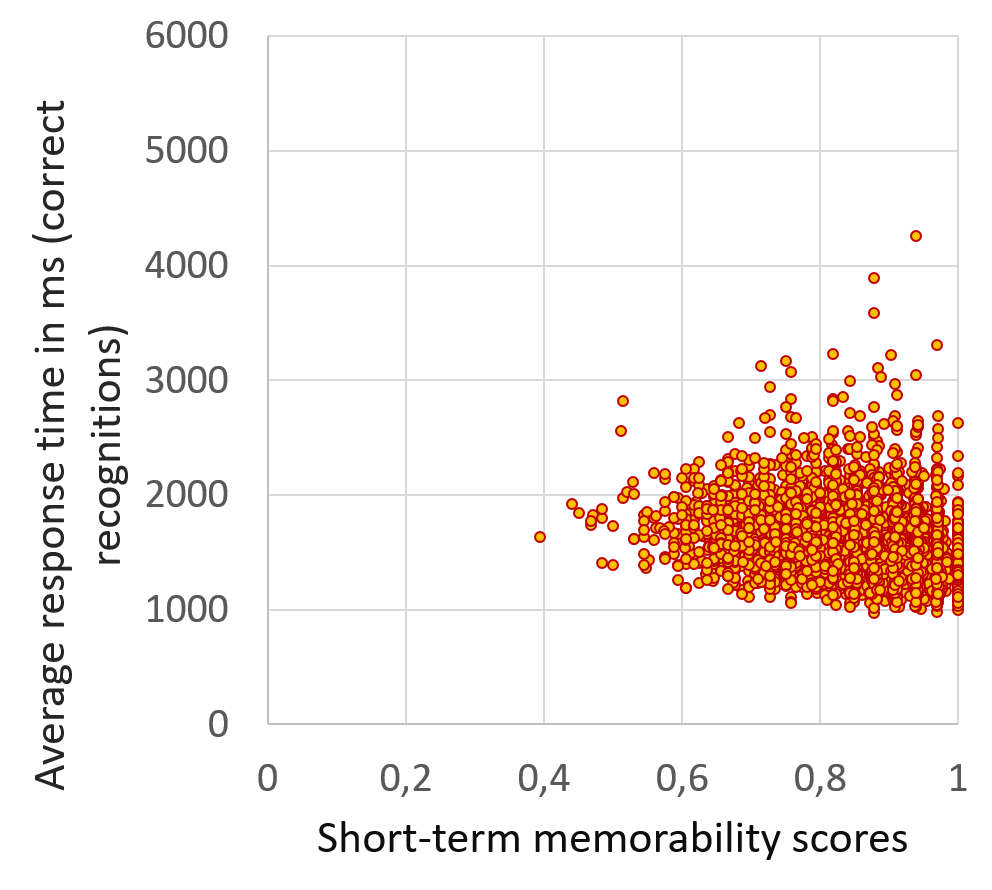}
	\includegraphics[width=0.42\columnwidth]{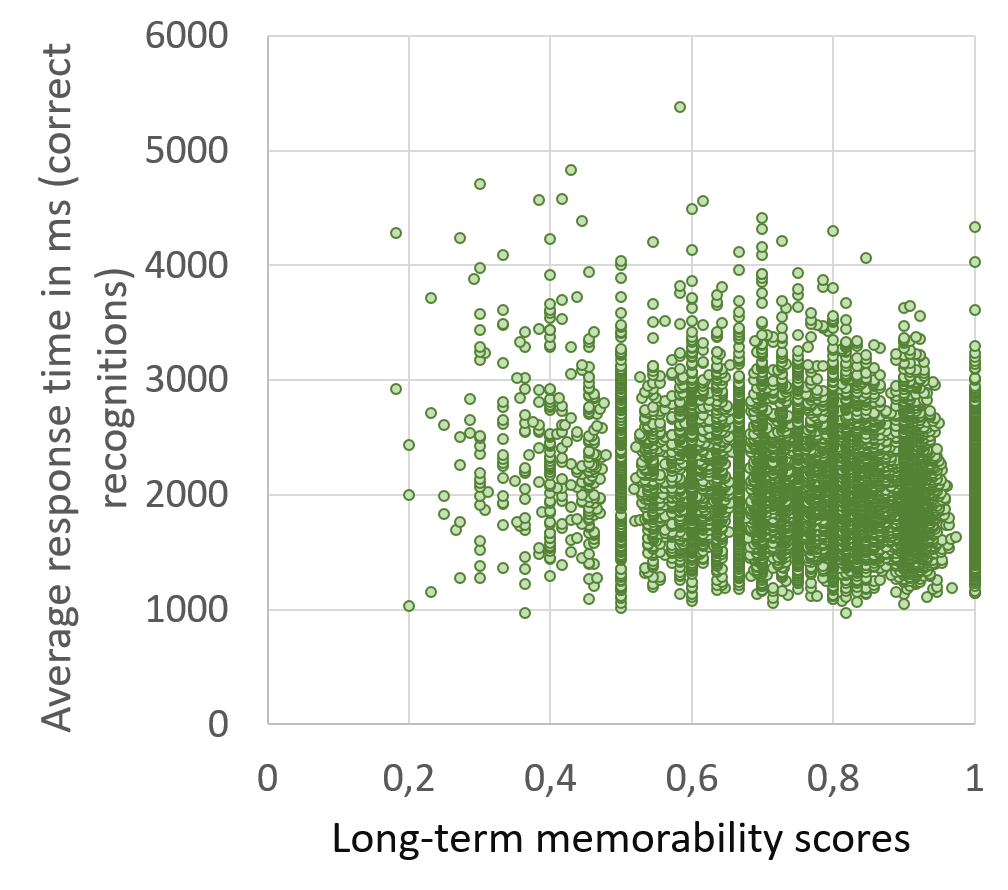}	\caption{\label{fig:response_time_vs_memorability}Average response time (correct recognitions only) as a function of memorability scores, for short-term (left) and long-term memorability (right).}
\end{figure}

We observed negative Pearson correlations between the mean response time to correctly recognize targets and their memorability scores, both for short-term ($r=0.307$, $p<.0001$) and long-term ($0.176$, $p<.0001$) memorability, as also illustrated in Fig. \ref{fig:response_time_vs_memorability}. 
This tends to prove that, globally, participants tended to answer more quickly for the most memorable videos than for the less memorable ones. 
This is consistent with \cite{cohendet_2018_annotating}, where the authors propose two explanations to this result: either the most memorable videos are also the most accessible in memory, and/or the most memorable videos contain more early recognizable elements than the less memorable ones.
As videos in VideoMem consist of semantic units with often one unique shot -- with most of the information already present from the beginning -- the first explanation sounds more suitable here.
This also suggests that participants tend to quickly answer after recognizing a repeated video (even though they did not receive any instruction to do so), maybe afraid of missing the time to answer, or to alleviate their mental charge.
This correlation highlights that the average response time might be a useful feature to further infer VM in computational models. 

The correlation is, however, lower for long-term memorability. One explanation might be that, after one day, remembering is more difficult. In connection with this explanation, we observed a significant difference between the mean response time to correctly recognize a video during step \#1 and during step \#2 ($1.43 sec.$ \textit{vs.} $3.37 sec.$), as showed by a Student's t-test ($t(9999)=-122.59, p<0001$).
Note that the Pearson correlation ($0.291$) between average response time per video for short-term and long-term memorability is close to the Pearson correlation ($0.329$) observed between short-term and long-term memorability scores (see Fig. \ref{fig:scatter_plot_st_vs_lt}, right).
Note that the mean response time for a false alarm was $3.17 sec.$ for step \#1 and $3.53 sec.$ for step \#2.

%figure Distribution of memorability scores + Mean response time vs. degrees of memorability

% ================================================================================================================================================== %
% ================================================================================================================================================== %
% ---------------------------------------------------------------------------------------------- %
%% PREDICTING VIDEO MEMORABILITY
% ---------------------------------------------------------------------------------------------- %
\section{Predicting video memorability}
\label{sec:predicting_memorability}

In this section we focus on predicting VM using various machine learning approaches. We pose the VM score prediction as a standard regression problem. 
Among the five models we proposed, the first two are IM models re-used as is on our data (see Section \ref{subsec:IM_models}). They will serve as a baseline for the purpose of performance comparison and understanding the correlation between IM and VM. We then propose in section \ref{subsec:IC_basedmodel} a simple model based on Image Captioning (IC) features, that derives from the finding in \cite{squalli_2018_deep, cohendet_2018_annotating}. 
The fourth model consists of a fine-tuned version of a state-of-the-art high performance DNN for the task of image recognition (Section \ref{subsec:ResNet}). We pursue our model investigation with a fine-tuning of an advanced complete IC model in section \ref{subsec:advanced_model}. 
In section \ref{ssec:results}, we analyze the prediction results of these five models. Last, in section \ref{subsec:attention_mechanism}, we modify the advanced IC-based model by adding an attention mechanism that helps us better understand what makes a content memorable.
Note that, for training (when applied) and evaluating the considered models, we split VideoMem dataset into training (6500 videos), validation (1500 videos), and test (2000 videos) sets, where the test set contains 500 videos having a greater number of annotations. Similarly to previous work in IM and VM, the prediction performance is evaluated in term of the Spearman's rank correlation between the ground truth and the predicted memorability scores.

\begin{figure}[!htbp]
	\centering
	\includegraphics[width=\columnwidth]{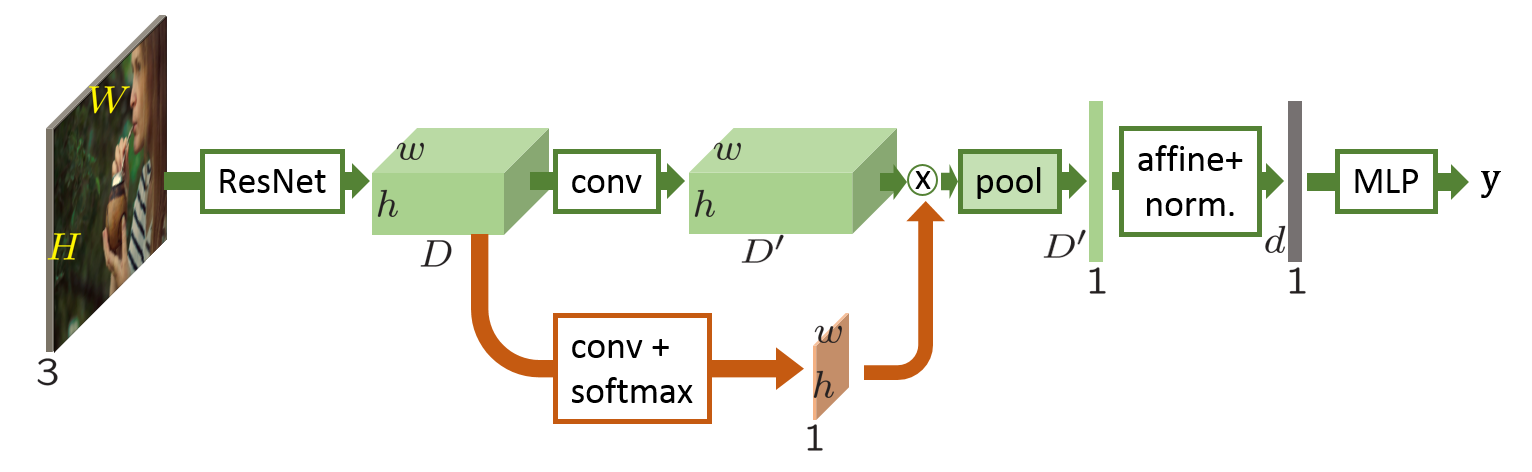}
	\caption{Semantic embedding based model without (green pipeline) and with an attention mechanism (adding the orange branch).}
	\label{fig:attention_model}
\end{figure}
% ============== %
\subsection{Image memorability based models}
\label{subsec:IM_models}

In order to investigate the correlation between IM and VM, we used two state-of-the-art models available for IM prediction to directly compute a memorability score for 7 successive frames in the video (one per second). The first IM model, so-called MemNet \cite{khosla_2015_understanding}, fine-tuned CNN pre-trained on both the ImageNet and Places databases using the LaMem dataset \cite{isola_2014_makes} and obtained prediction results close to human consistency on LaMem. The second considered IM model \cite{squalli_2018_deep} obtained even better performance on LaMem. In this model, the authors used the pre-trained VGG16 network for their CNN feature and a pre-trained IC model as an extractor for a high level visual semantic feature. 
Finally, these two features are combined and MLP is used as classifier. For both models, the final VM score is then obtained by averaging the 7 frame scores.

\subsection{Image captioning-based model}
\label{subsec:IC_basedmodel}

As scene semantic features derived from an image captioning system \cite{KirosSZ14} have been shown to well characterize the memorability of images \cite{squalli_2018_deep} and videos \cite{cohendet_2018_annotating}, we also investigated the use of an IC system as feature extractor. This system builds an encoder comprising a CNN and a long short-term memory recurrent network (LSTM) for learning a joint image-text embedding. We extracted the projected CNN feature (of dimension 1024) in such joint 2D embedding space, for each of the three (first, middle, last) frames from each video. The 3 frame features are then concatenated and given as input to a MLP with mean square error (MSE) measure as regression loss and trained to predict the IM score. 
Final result is obtained with MLP parameters: one hidden layer with 1500 neurons, optimizer=IBLGS, activation=tanh, learning rate (lr)=1e-3. 

% ============== %
\subsection{Fine-tuned ResNet model}
\label{subsec:ResNet}
Instead of using a fix feature extractor as investigated in the previous model, we also fine-tuned the state-of-the-art ResNet models designed for the image recognition task \cite{he_2016_resnet}. For this, we replaced the last fully connected layer of ResNet by a new one dedicated to our considered regression task. This last layer was first trained alone for 5 epochs (Adam optimizer, batchsize=32, lr=1e-3), then the whole network was re-trained for more epochs (same parameters, but lr=1e-5). Input data was once again all 7 frames of each video (one per second, each frame being assigned the same ground-truth score as the video) from VideoMem, mixed with images from LaMem, to enlarge the size of the overall training dataset. For the latter images, we normalized the ground-truth scores to be in the same range as those of VideoMem. Some data augmentation was conducted: random center cropping of 224x224 after resizing of the original images and horizontal flip, followed by a mean normalization computed on ImageNet. We fine-tuned two variants of ResNet: ResNet18 and ResNet101 and found that the latter gave best performance on the validation set. Note that, for ResNet101, the model was fine-tuned only to predict short-term memorability as LaMem dataset contains only short-term scores.

%Table: Perf. of the different models
\begin{table*}[ht]
	\centering
	%\small
	\begin{tabular}{ |c|c|c|c|c|c|c|c|c|c|c| }
		\hline
		\multirow{2}{*}{Models} & \multicolumn{3}{|c|}{short-term memorability} & \multicolumn{3}{|c|}{long-term memorability}  \\
		\cline{2-7}
		& \color{gray}{validation} & test & test (500) & \color{gray}{validation} & test & test (500) \\ \hline
		MemNet (Sec. \ref{subsec:IM_models}) & \color{gray}{ 0.397 } & 0.385 & 0.426 & \color{gray}{ 0.195 } & 0.168 & 0.213 \\ \hline
		Squalli \textit{et al.} (Sec. \ref{subsec:IM_models}) & \color{gray}{ 0.401 } & 0.398 & 0.424 & \color{gray}{ 0.201 } & 0.182 & 0.232 \\ \hline
		IC-based model (Sec. \ref{subsec:IC_basedmodel}) & \color{gray}{0.492} & 0.442 & 0.514 & \color{gray}{0.22} & 0.201 & 0.188 \\ \hline%IC feature alone
		ResNet101 (Sec. \ref{subsec:ResNet}) & \color{gray}{0.498} & 0.46 & 0.527 & \color{gray}{0.222} & 0.218 & 0.219 \\ \hline
		Semantic embedding model (Sec. \ref{subsec:advanced_model})& \color{gray}{\bf{0.503}} & \bf{0.494} & \bf{0.565} & \color{gray}{\bf{0.26}} & \bf{0.256} & \bf{0.275} \\ \hline
	\end{tabular}
	\caption{Results in terms of Spearman's rank correlation between predicted and ground truth memorability scores, on the validation and test sets, and on the 500 most annotated videos of the dataset (test (500)) that were placed in the test set.}
	\label{tab:results}
\end{table*}
% ================================================================================================================================================ %
\subsection{Semantic embedding based model}
\label{subsec:advanced_model}

Following the idea of model fine-tuning and with an attempt of benefiting from the performance of IC features, we used a state-of-art visual semantic embedding pipeline used for image captioning \cite{engilberge_2019}, on top of which a 2-layer MLP is added, to regress the feature space to a single memorability score. The overall architecture is shown in Fig. \ref{fig:attention_model}, in the green pipeline.
Similarly to the model described in section \ref{subsec:ResNet}, this model first predicts memorability scores for 7 successive frames (one per second), then the final prediction at video level being computed by averaging those 7 values. It is also fine tuned on VideoMem and LaMem data for short-term only.
The training is done using the Adam optimizer and is divided in two steps: in the first 10 epochs only the weights of the MLP are updated while those of the IC feature extractor remain frozen. Later the whole model is fine-tuned. The learning rate is initialized to 0.001 and divided in half every three epochs. It is important to note that the original IC model was trained with a new ranking loss (\emph{i.e., } Spearman surrogate) proposed in \cite{engilberge_2019}. This new loss has proved to be highly efficient for ranking tasks as claimed in \cite{engilberge_2019}. 
For the fine-tuning however, the training starts with a $\ell_1$ loss as initialization step, before coming back to the ranking loss. The original model was indeed trained for scores in [-1;1], while our memorability scores are in [0;1]. The $\ell_1$ loss forces the model to adapt to this new range.

\begin{figure}[!htbp]
	\centering
	\includegraphics[width=0.99\columnwidth]{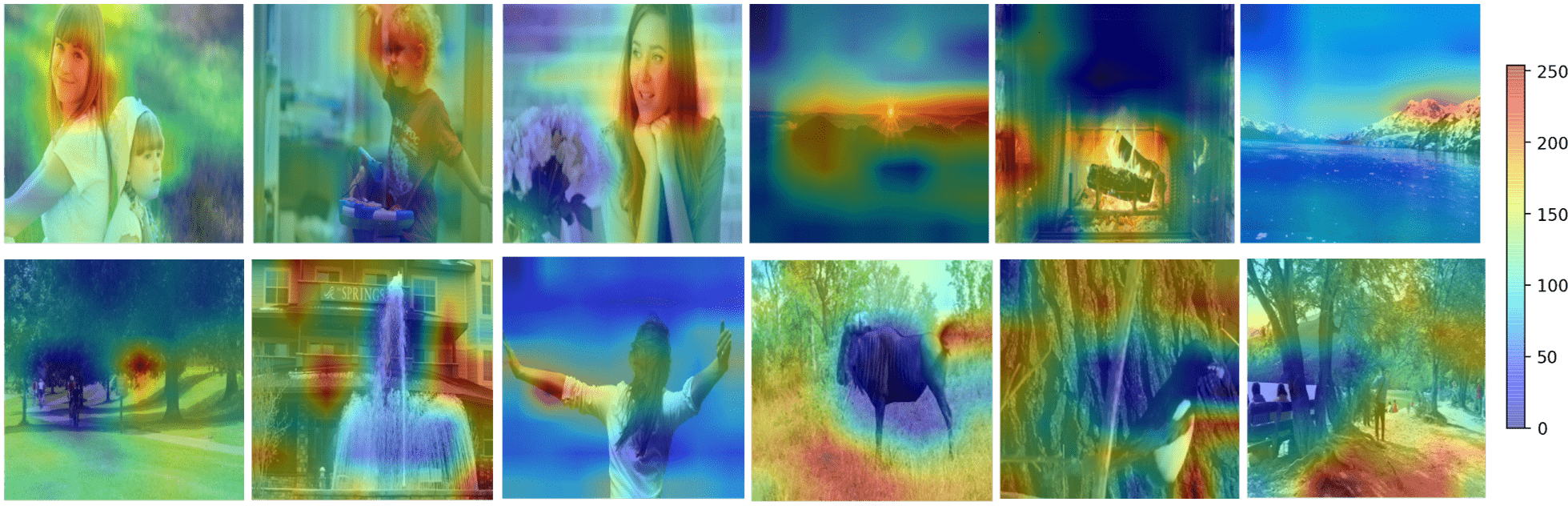}
	\caption{Visualization of the attention mechanism's output. The model focuses either on close enough faces or main objects when the image is mostly empty or black (row \#1), or it focuses on details outside the main objects (row \#2).}
	\label{fig:attention_results}
\end{figure}

% ================================================================================================================================================= %
\subsection{Prediction results}
\label{ssec:results}

From the results in Table \ref{tab:results}, we may draw several conclusions.
First, it is possible to achieve already quite good results in VM prediction using models designed for IM prediction. This means that the memorability of a video is correlated to some extent with the memorability of its constituent frames.
In accordance with the literature, the model of \cite{squalli_2018_deep} performed a little better than the model of \cite{khosla_2015_understanding} again for VM prediction. Also, all other models, dedicated to the task, show significantly better performances than the baselines. Their ranking confirms what was expected: IC features performed slightly worse than the two complete fine-tuned models, but among those two, the fine-tuned IC model is the best, as it leverages both the dedicated fine-tuning and the use of high level semantic information.
For all models, we note that performances were lower for long-term memorability.
One interpretation might be that the memorability scores for long-term are based on a smaller number of annotations than for short-term, so they probably capture a smaller part of the intrinsic memorability. However, it may also highlight the difference between short-term and long-term memorability, the latter being more difficult to predict as it is more subjective, while both being still -- though not perfectly -- correlated.
The performances of our models on the 500 most annotated videos are better.
This reveals that our dataset might benefit from a larger number of annotations. Last, compared to annotation consistency values, performances remain lower, showing that there is still room for improvement.

% ================================================================================================================================================= %
\subsection{Intra-memorability visualization}
\label{subsec:attention_mechanism}%to write after others parts
%Here we put images from attention model

To better understand what makes an image memorable, we added an attention mechanism to our best model. It will then learn what regions in each image contribute more to the prediction. For this purpose, a convolutional layer is added in parallel with the last convolutional layer of the feature extractor part. It outputs a 2D attention map which goes through a softmax layer and is multiplied with the last convolution map of the visual pipeline as shown in Fig. \ref{fig:attention_model} (orange branch). Note that for training we used only the $\ell_1$ loss, and the other same parameters.
An empirical study of the resulting attention maps tends to separate them in two categories. In the first one, when image frames contain roughly one main object and no or rare information apart from this main object (this might be because the background is dark or uniform), it seems that the model focuses, as expected intuitively, on the main object and even, in the case of large enough faces, on details of the faces, as if trying to remember the  specific features of faces. Example results for images in the first category can be found in Fig. \ref{fig:attention_results}, first row.
In the second category that groups all other frames, with several main and secondary objects, cluttered background, etc., it seems on the contrary that the model focuses on all but the main objects/subjects of the images, as if trying to remember little details that will help it differentiate the image from another similar one. Or said differently, the second category shows results that might be interpreted as a second memorization process, once the first one -- focusing on the main object -- is already achieved. Examples for the second category can be found in the second row of Fig. \ref{fig:attention_results}.

% ---------------------------------------------------------------------------------------------- %
%% CONCLUSION
% ---------------------------------------------------------------------------------------------- %
\section{Conclusions}
\label{sec:conclusions}

In this work, we presented a novel memory game based protocol to build VideoMem, a premier large-scale VM dataset. Through an in-depth analysis of the dataset, we highlighted several important factors concerning the understanding of VM: human \emph{vs.} annotation consistency, memorability over time, and memorability \emph{vs.} response time. We then investigated several baselines and advanced DNN models for VM prediction. 
Our proposed model with \emph{spatial} attention mechanism allows to visualize, and thus better understand what type of visual content is more memorable. Future work would be devoted to improve results for both short-term and long-term memorability with a focus on temporal aspects of the video, \emph{e.g.} by adding \emph{temporal} attention model and recurrent neural network blocks to the workflow.

% ---------------------------------------------------------------------------------------------- %
%% REFERENCES
% ---------------------------------------------------------------------------------------------- %
{\small
\bibliographystyle{ieee}
\bibliography{references}

\begin{thebibliography}{10}\itemsep=-1pt

\bibitem{engilberge_2019}
Authors.
\newblock Sodeep: a sorting deep net to learn ranking loss surrogates.
\newblock CVPR 2019 Submission ID 5921, Supplied as additional Material
  cvpr5921.pdf, 2019.

\bibitem{baveye_2016_deep}
Y.~Baveye, R.~Cohendet, M.~Perreira Da~Silva, and P.~Le~Callet.
\newblock Deep learning for image memorability prediction: the emotional bias.
\newblock In {\em Proc. ACM Int. Conf. on Multimedia (ACMM)}, pages 491--495,
  2016.

\bibitem{brady_2008_visual}
T.~F. Brady, T.~Konkle, G.~A. Alvarez, and A.~Oliva.
\newblock Visual long-term memory has a massive storage capacity for object
  details.
\newblock {\em Proceedings of the National Academy of Sciences},
  105(38):14325--14329, 2008.

\bibitem{celikkale_2013_visual}
B.~Celikkale, A.~Erdem, and E.~Erdem.
\newblock Visual attention-driven spatial pooling for image memorability.
\newblock In {\em Proc. IEEE Int. Conf. on Computer Vision and Pattern
  Recognition Workshops}, pages 976--983, 2013.

\bibitem{cohendet_2016_prediction}
R.~Cohendet.
\newblock {\em Pr{\'e}diction computationnelle de la m{\'e}morabilit{\'e} des
  images: vers une int{\'e}gration des informations extrins{\`e}ques et
  {\'e}motionnelles}.
\newblock PhD thesis, Nantes, 2016.

\bibitem{cohendet_2018_mediaeval}
R.~Cohendet, C.-H. Demarty, N.~Q.~K. Duong, M.~Sj{\"o}berg, B.~Ionescu, and
  T.-T. Do.
\newblock Mediaeval 2018: Predicting media memorability task.
\newblock In {\em Proc. of the MediaEval Workshop}, 2018.

\bibitem{cohendet_2016_using}
R.~Cohendet, A.-L. Gilet, M.~P. Da~Silva, and P.~Le~Callet.
\newblock Using individual data to characterize emotional user experience and
  its memorability: Focus on gender factor.
\newblock In {\em Proc. Int. Conf. on Quality of Multimedia Experience
  (QoMEX)}, pages 1--6. IEEE, 2016.

\bibitem{cohendet_2018_annotating}
R.~Cohendet, K.~Yadati, N.~Q.~K. Duong, and C.-H. Demarty.
\newblock Annotating, understanding, and predicting long-term video
  memorability.
\newblock In {\em Proc. of the ICMR 2018 Workshop, Yokohama, Japan, June
  11-14}, 2018.

\bibitem{cowan_1998_attention}
N.~Cowan.
\newblock {\em Attention and memory: An integrated framework}.
\newblock Oxford University Press, 1998.

\bibitem{ebbinghaus_memory:_1913}
H.~Ebbinghaus.
\newblock {\em Memory; a contribution to experimental psychology}.
\newblock New York city, Teachers college, Columbia university, 1913.

\bibitem{ebbinghaus_1913_memory}
H.~Ebbinghaus.
\newblock {\em Memory: A contribution to experimental psychology}.
\newblock Number~3. University Microfilms, 1913.

\bibitem{fajtl_2018_amnet}
J.~Fajtl, V.~Argyriou, D.~Monekosso, and P.~Remagnino.
\newblock Amnet: Memorability estimation with attention.
\newblock In {\em Proceedings of the IEEE Conference on Computer Vision and
  Pattern Recognition}, pages 6363--6372, 2018.

\bibitem{furman_2007_they}
O.~Furman, N.~Dorfman, U.~Hasson, L.~Davachi, and Y.~Dudai.
\newblock They saw a movie: long-term memory for an extended audiovisual
  narrative.
\newblock {\em Learning \& memory}, 14(6):457--467, 2007.

\bibitem{han_2015_learning}
J.~Han, C.~Chen, L.~Shao, X.~Hu, J.~Han, and T.~Liu.
\newblock Learning computational models of video memorability from fmri brain
  imaging.
\newblock {\em IEEE transactions on cybernetics}, 45(8):1692--1703, 2015.

\bibitem{he_2016_resnet}
K.~He, X.~Zhang, S.~Ren, and J.~Sun.
\newblock Deep residual learning for image recognition.
\newblock In {\em Proc. IEEE Int. Conf. on Computer Vision and Pattern
  Recognition (CVPR)}. IEEE, 2016.

\bibitem{isola_2014_makes}
P.~Isola, J.~Xiao, D.~Parikh, A.~Torralba, and A.~Oliva.
\newblock What makes a photograph memorable?
\newblock {\em IEEE Transactions on Pattern Analysis and Machine Intelligence},
  36(7):1469--1482, 2014.

\bibitem{isola_2011_makes}
P.~Isola, J.~Xiao, A.~Torralba, and A.~Oliva.
\newblock What makes an image memorable?
\newblock In {\em Proc. IEEE Int. Conf. on Computer Vision and Pattern
  Recognition (CVPR)}, pages 145--152. IEEE, 2011.

\bibitem{jing_2017_predicting}
P.~Jing, Y.~Su, L.~Nie, and H.~Gu.
\newblock Predicting image memorability through adaptive transfer learning from
  external sources.
\newblock {\em IEEE Transactions on Multimedia}, 19(5):1050--1062, 2017.

\bibitem{kensinger_2008_memory}
E.~A. Kensinger and D.~L. Schacter.
\newblock Memory and emotion.
\newblock {\em Handbook of emotions}, 3:601--617, 2008.

\bibitem{khosla_2015_understanding}
A.~Khosla, A.~S. Raju, A.~Torralba, and A.~Oliva.
\newblock Understanding and predicting image memorability at a large scale.
\newblock In {\em Proc. IEEE Int. Conf. on Computer Vision (ICCV)}, pages
  2390--2398, 2015.

\bibitem{kim_2013_relative}
J.~Kim, S.~Yoon, and V.~Pavlovic.
\newblock Relative spatial features for image memorability.
\newblock In {\em Proceedings of the 21st ACM international conference on
  Multimedia}, pages 761--764. ACM, 2013.

\bibitem{KirosSZ14}
R.~Kiros, R.~Salakhutdinov, and R.~S. Zemel.
\newblock Unifying visual-semantic embeddings with multimodal neural language
  models.
\newblock {\em CoRR}, abs/1411.2539, 2014.

\bibitem{lahrache_2016_bag}
S.~Lahrache, R.~El~Ouazzani, and A.~El~Qadi.
\newblock Bag-of-features for image memorability evaluation.
\newblock {\em IET Computer Vision}, 10(6):577--584, 2016.

\bibitem{mancas_2013_memorability}
M.~Mancas and O.~Le~Meur.
\newblock Memorability of natural scenes: The role of attention.
\newblock In {\em Proc. IEEE Int. Conf. on Image Processing (ICIP)}, pages
  196--200, 2013.

\bibitem{mcgaugh_2000_memorycentury}
J.~L. McGaugh.
\newblock Memory--a century of consolidation.
\newblock {\em Science}, 287(5451):248--251, 2000.

\bibitem{nadel_1997_memory}
L.~Nadel and M.~Moscovitch.
\newblock Memory consolidation, retrograde amnesia and the hippocampal complex.
\newblock {\em Current opinion in neurobiology}, 7(2):217--227, 1997.

\bibitem{quillan_1966_semantic}
M.~R. Quillan.
\newblock Semantic memory.
\newblock Technical report, Bolt Beranek and Newman Inc Cambridge MA, 1966.

\bibitem{revlin_cognition:_2012}
R.~Revlin.
\newblock {\em Cognition: {Theory} and {Practice}}.
\newblock Palgrave Macmillan, July 2012.

\bibitem{richardson_1988_measures}
A.~Richardson-Klavehn and R.~A. Bjork.
\newblock Measures of memory.
\newblock {\em Annual review of psychology}, 39(1):475--543, 1988.

\bibitem{shekhar_2017_show}
S.~Shekhar, D.~Singal, H.~Singh, M.~Kedia, and A.~Shetty.
\newblock Show and recall: Learning what makes videos memorable.
\newblock In {\em Proceedings of the IEEE Conference on Computer Vision and
  Pattern Recognition}, pages 2730--2739, 2017.

\bibitem{squalli_2018_deep}
H.~Squalli-Houssaini, N.~Q. Duong, M.~Gwena{\"e}lle, and C.-H. Demarty.
\newblock Deep learning for predicting image memorability.
\newblock In {\em 2018 IEEE International Conference on Acoustics, Speech and
  Signal Processing (ICASSP)}, pages 2371--2375. IEEE, 2018.

\bibitem{standing_1973_learning}
L.~Standing.
\newblock Learning 10000 pictures.
\newblock {\em Quarterly Journal of Experimental Psychology}, 25(2):207--222,
  1973.

\bibitem{zarezadeh_2017_image}
S.~Zarezadeh, M.~Rezaeian, and M.~T. Sadeghi.
\newblock Image memorability prediction using deep features.
\newblock In {\em Electrical Engineering (ICEE), 2017 Iranian Conference on},
  pages 2176--2181. IEEE, 2017.

\end{thebibliography}
}

\end{document}